# Concurrent Activity Recognition with Multimodal CNN-LSTM Structure

Xinyu Li[1], Yanyi Zhang[1], Jianyu Zhang[1], Shuhong Chen[1], Ivan Marsic[1],
Richard A. Farneth[2], Randall S. Burd[2]

1 Department of Electrical and Computer Engineering, Rutgers University, New Brunswick, NJ, USA
{xinyu.li1118, yz593, jz549, sc1624, marsic}@rutgers.edu
2 Division of Trauma and Burn Surgery, Children's National Medical Center, Washington, DC, USA
{rfarneth, rburd}@childrensnational.org

## ABSTRACT

We introduce a system that recognizes concurrent activities from real-world data captured by multiple sensors of different types. The recognition is achieved in two steps. First, we extract spatial and temporal features from the multimodal data. We feed each datatype into a convolutional neural network that extracts spatial features, followed by a long-short term memory network that extracts temporal information in the sensory data. The extracted features are then fused for decision making in the second step. Second, we achieve concurrent activity recognition with a single classifier that encodes a binary output vector in which elements indicate whether the corresponding activity types are currently in progress. We tested our system with three datasets from different domains recorded using different sensors and achieved performance comparable to existing systems designed specifically for those domains. Our system is the first to address the concurrent activity recognition with multisensory data using a single model, which is scalable, simple to train and easy to deploy.

## CCS Concepts

I.5.2 [Pattern Recognition]: Design Methodology–Classifier design & evaluation; C.3 [Special-Purpose and Application-Based Systems]: Real-time and embedded systems.

## 1. INTRODUCTION

We present a neural network system for concurrent activity recognition from multisensory data. We designed and concatenated the modules (preprocessor, feature extractor, and classifier) to achieve high recognition accuracy while maintaining a low hardware resources requirement. Activity recognition has been researched for decades with a variety of sensors. While each sensory type has advantages in certain applications [1][2], it still faces limitations. Recent research on multisensory systems exploits the advantages of different datatypes. For example, mobile sensors such as wearable devices and RFIDs may be used together [1][3]. However, researchers often propose different features and classifiers, designed for specific scenarios. Because it is difficult to measure the effectiveness of features and classifier across different applications, manual feature selection is often arbitrary and lacks generalizability. Instead of using arbitrary manufactured features, we implemented automatic feature extraction using deep convolutional neural networks (ConvNet), based on successful previous implementations [3][4]. However, some sensory data, such as those obtained from passive RFID tags using multiple reader antennas, need to be properly arranged to maintain spatial and temporal relations in the input space before feeding them into a ConvNet for feature extraction.

Unlike visual shapes or spoken words, activities are abstract concepts, so both spatial and temporal information must be considered. Previous research often treated activity recognition as continuous "image classification" [5] or "sequential data classification" [6] problems, ignoring important temporal or spatial associations in the data. Attempts at spatio-temporal feature extraction include the usage of manufactured spatial features and fixed time windows for temporal associations. Such manufactured features do not generalize well and fixed time windows do not scale well for activities of different durations that commonly occur in real-world applications. ConvNets have been widely used for spatial feature extraction, and long-short term memory nets (LSTM) have been successfully implemented for sequential data modeling [7], we used them together for spatio-temporal feature extraction. Unlike existing models for image captioning or image description that use ConvNet and LSTM in parallel [8], our ConvNet and LSTM are connected in series to learn temporal associations between spatial features, similar to [9].

We address the challenge of recognizing concurrent activities, which are common in real-world scenarios. Most activity recognition research has focused on single-person or non-concurrent team activity recognition. Recognition of individual activities is a multiclass classification problem that can be solved using a multiclass classifier. Recognition of concurrent activities has been attempted using multiple binary classifiers [10], but such systems do not scale to large number of activities. We introduce an *encoder framework* that outputs a binary code as a prediction, where each bit denotes the status of an activity (one for "in progress" and zero for "not in progress").

Our CNN and LSTM structure for spatio-temporal feature extraction can accept input from different sensors, making it generalizable to many activity recognition applications. We tested our system with three datasets recorded using different sensors in different domains:

1. Dataset containing 35 activity types recorded during 42 actual trauma resuscitations. The passive RFID system, depth camera and microphone array were used for data collection. Our experimental results showed 93.48% overall accuracy with 0.32 mean average precision (mAP), outperforming existing systems applied in a similar environment. Trauma resuscitation is one of the most challenging environments for activity recognition because 10 or more providers are crowded around the patient bed and engaged in fast-paced work. Additional challenges stem from the depth camera's low resolution, frequent view occlusion, and RFID radio signal interferences caused by a dynamic and crowded setting.

2. The Charades dataset [11] of 9,848 videos of indoors daily activities performed by a single person (another person may be in the scene). This dataset with 157 activity classes including concurrent activities was collected using a single video camera and labeled with Amazon Mechanical Turk. Our experimental results showed concurrent activity recognition on a large-scale dataset with a performance competitive with existing systems.

3. Olympic sports dataset with 16 sports activities performed by athletes, without concurrent activities [12]. Our system achieved performance comparable to existing systems that used the same dataset. This experiment demonstrated that our system can also perform single-activity recognition although it treats the classification as a multiclass coding problem.

The contributions of this paper are:

1. A multimodal CNN-LSTM structure for extracting spatio-temporal features from multisensory, multimodal data for concurrent activity recognition.

2. A coding network layer that can be trained with backpropagation to encode a binary output vector in which elements indicate whether the corresponding activity types are currently in progress.

3. The memory usage analysis and evaluation of the system with multiple real-world datasets that can be used as a reference by other researchers.

The rest of this paper is organized as follows. Section 2 reviews related work on activity recognition and deep learning. Section 3 introduces the challenges of concurrent activity recognition and Section 4 describes our proposed system framework and Section 5 details the system implementation. Section 6 reports the experimental results and our analysis of these results. Section 7 discusses the results and Section 8 concludes the paper.

## 2. RELATED WORK

Activity recognition has been studied for decades in various application scenarios using different sensor types. The early research trained shallow classifiers on staged activities collected by a single sensor [13][14]. Though they proved applicable, the early single-sensor based systems were constrained by sensor specific limitations. For example, the glove with RFID antenna is bulky for real-world applications and the system could only detect activities that use [14]. In general, data from individual sensors are inherently insufficient for complex activity recognition.

Multisensory systems have been proposed to address these challenges. Sensors in different locations or different sensor types provide additional information for activity recognition [15] [16]. Our system is able to work with multiple sensor types, including depth video, audio, and RFID.

With sufficient multisensory data, the common approach was to utilize different sensor data for different activity recognition [17]. A step further, the hierarchical model was proposed to first predict low level activities (such as the human-object interaction) and further predict the high-level meaningful activity status based low-level information [10]. The problem is, the hierarchical model still relies on manufactured features and shallow classifiers, which often do not work well in real world scenarios with large number of activities. In addition, the hierarchical model is an error propagation system which often heavily relies on the low-level activity detection.

More recently, following the success of deep learning at image classification [18] and image description [6], researchers implemented deep learning for sensor-based activity recognition [19][20]. For example, the convolutional neural network structure, which does not require manufactured features, has been used to extract features from RFID data collected by multiple antennas in a medical setting [3]. Based on these successful passive RFID implementations [3], in our system, we used a deep ConvNet structure to learn the meaningful features and a multimodal structure previously proposed for visual-acoustic input [21][22] to handle multisensory data. To learn both the temporal and spatial associations in the data, we supersede the ConvNet with a long-short term memory networks [9]. After the ConvNet, which is commonly used for learning sequential associations.

Concurrent activities, which are common in real-world scenarios, have not been addressed in activity recognition research until recently [23][24]. We present a more efficient and accurate approach to concurrent activity recognition.

## 3. PROBLEM DESCRIPTION

We identified the following challenges for activity recognition in real-world processes:

1. **Concurrent activities are common**: With $N$ activities, there are $N$ labels for individual-activity prediction but $2^N$ potential combinations of concurrent activities. An efficient classifier structure is needed for concurrent activity prediction. Our analysis of the 42 trauma resuscitations dataset showed that more than 50% of time instances had at least two concurrent activities (Fig. 1(a)).

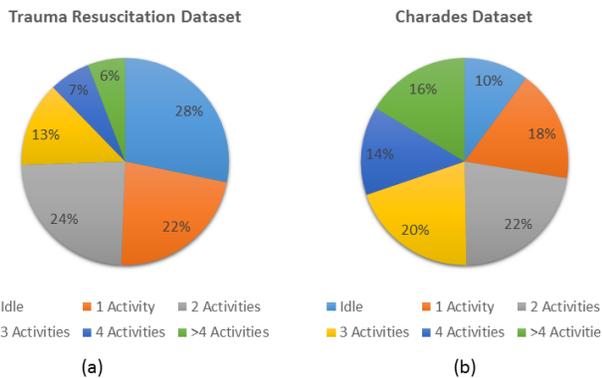
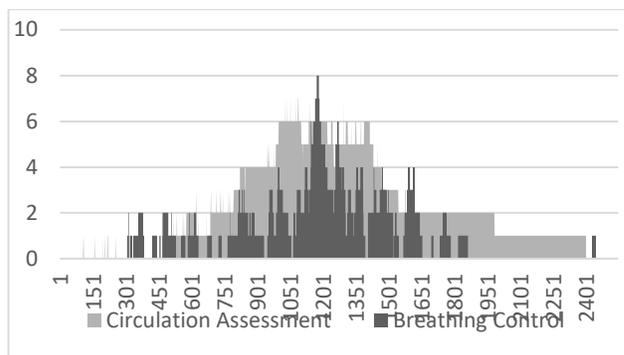

**Fig. 1.** Distribution of concurrent activities in one-second time intervals for two datasets used in this study.

**Fig. 2.** Co-occurrence of Breathing Control and Circulation Assessment over time in 42 trauma resuscitations.

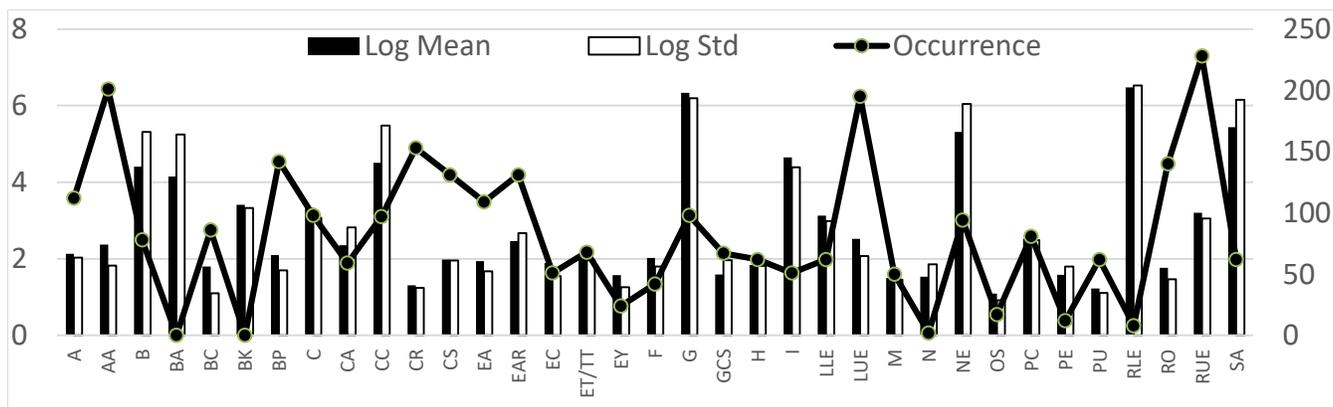

**Fig. 3.** The log duration and standard deviation of 35 activities in our trauma resuscitation dataset (histogram with units on the left axis) and the # of occurrence of each activity in the dataset (solid line with units on the right axis).

Even for daily living scenarios (Charades dataset) there were more than 70% time instances with at least two concurrent activities (Fig. 1(b)).

2. **Multimodal features**: Different features are best suited to recognize different activities. With multiple sensors and a large number of activities, manual feature selection is neither scalable nor generalizable.

3. **Temporal associations between sensors and between concurrent activities**: Human body posture changes during activities. The environmental sound changes when certain equipment is in use and radio signals change when tagged objects are in use. Concurrent activities also exhibit temporal associations. For example, although resuscitations have different durations, many activities are likely to be performed simultaneously (Fig. 2).

4. **Diverse activity durations and variable activity frequency**: The same activity can be performed faster or slower. The 35 activity types performed in 42 trauma resuscitations had very different duration (Fig. 3). Using a fixed time window would not work well with diverse durations of activities. In addition, some activities occur more often than others (Fig. 3), meaning that some activities will be poorly represented in the training data.

5. **Ignored activities and idle intervals**: Some work activities may not be of interest and may be excluded from observation. The activity recognition system would be unaware of these excluded activities. With fewer tracked activities there will be fewer co-occurring activities or even some false "idle periods", resulting in fewer temporal associations among activities. These phenomena may impact the recognition performance.

We assume that at most one activity of each type can take place at any time instant.

## 4. SYSTEM STRUCTURE
### 4.1 Overall Structure

Our activity recognition system consists of four major modules connected in series (Fig. 4):

- **Data preprocessing** formats the sensory data into a ConvNet-ready representation. The ConvNet's input data is usually either a single feature map or several stacked ones. These maps should represent data redundancy and spatial relationships for the convolution and pooling layers to learn abstract features.

- **Feature extraction** from sensor data takes place in ConvNet structures that output vectors of features. Previous research successfully used ConvNets to extract features from images [5], audio [25], and RFID data [3]. We adopted a VGG-like structure [5] and the full convolutional network structure [26]. The output is a vector of features.

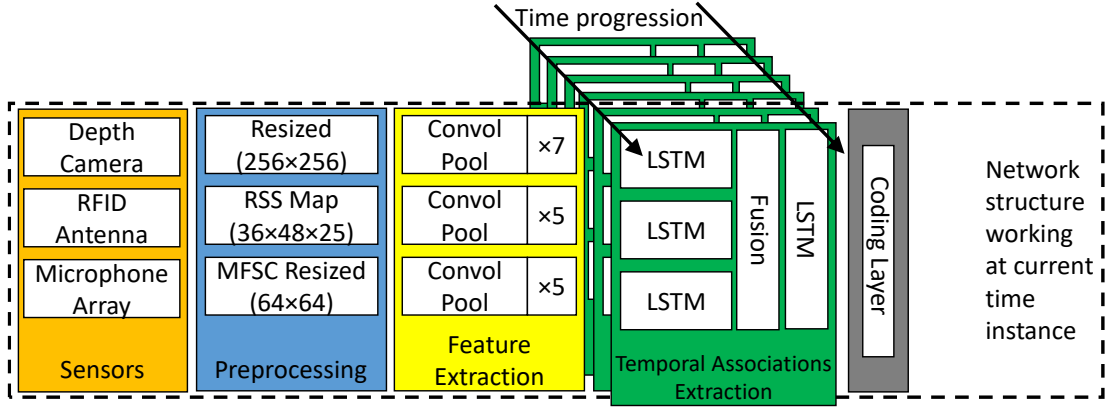

**Fig. 4.** Structure of our concurrent activity recognition system operating at a given time instance. The stacked part (green) symbolizes the LSTM memory evolution over time.

- **Temporal association extraction** using recurrent neural networks (RNNs) [27] because ConvNets alone are not suitable for temporally associated activities. Given that we are primarily interested in processes that last about an hour or less, we used the LSTM structure to model the time dependencies while avoiding exploding or vanishing gradients during long term training [28].
- **Decision-making** uses the extracted spatio-temporal features for concurrent activity prediction. Because of concurrency, the decision-making module cannot be a multiclass-classifier, like the common softmax layer. Given many activities, independent binary classifiers would require large computer memory, so we opted for fully-connected sigmoid activation layers to encode concurrent activity predictions.

These modules are described next.

### 4.2 Sensory Data Preprocessing

Our system can accept input from any type of sensor. We used an existing dataset [3] recorded using a depth camera, a microphone array, and a passive RFID system to demonstrate image, audio, and mobile sensor data preprocessing. A depth image is a 2D matrix, where each pixel value represents the distance between the camera and the imaged physical object. Since the depth images contain random hardware noise (undefined value pixels), we compensate the hardware noise using existing methods [29]. We also resized the depth images to 256×256px down from 512×424px. Other image types (RGB, RGBD) can be resized and fed into our ConvNet as well.

For each second of the audio data, we extracted time-frequency maps and fed them to the ConvNet. Mel-frequency cepstral coefficients (MFCCs) are commonly used for speech recognition with shallow classifiers [30] and deep neural networks. However, as argued previously [31], discrete cosine transformations (DCT) project the spectral energies onto new bases that may not maintain spatial locality. As suggested, we directly used the log-energy (MFSC) without performing DCT:

$$S(m) = \log\left(\sum_{k=0}^{N-1} |X(k)|^2 H_m(k)\right), \quad 0 \leq m \leq M$$

where $S(m)$ is the log-energy for band pass filter $m$, $X(k)$ denotes the Fourier transform of each audio frame of length $N$, and $H_m(k)$ is the Hamming window response at frequency $k$. We used 20 frames with 512 points per frame ($N$=512) for every second and 36 frequency bands ($M$=36). For multichannel audio recordings, the MFSC can be extracted from each individual channel.

Unlike image and audio, mobile sensors such as RFIDs have had relatively few deep learning implementations. Previous research preprocessed the data into a 3D antenna-object-time matrix [3]. Although feasible, this data format faces two problems: (a) there is no redundancy for ConvNet pooling operations, and (b) the spatial relationships between reader antennas and tags present in the received signal strength (RSS) data are not well represented. We introduce *RSS maps*, a new RFID data representation suitable for spatial feature extraction in ConvNets. An RSS map projects the RSSs received from each tag onto the effective field of coverage for each antenna (Fig. 5 right). In our experiments, 7 of the 8 reader antennas were hung on the ceiling facing down (black boxes in Fig. 5 left). We approximated antenna's coverage area by a circle on the room floorplan (scale: 1px ↔ 1dm$^2$). The circle radii were manually measured by moving a tag horizontally away from the antenna in several directions, always starting below the antenna moving away until the tag could not be detected by the antenna. The coverage radius was determined as the average tag-visibility-loss distance from these experiments. In our case, most tags were 0.6 meters above the ground (approximately at the height of person's hands, assuming that tagged objects are used for work), and were visible laterally up to about 1.2 meters away from an antenna. For the tilted antenna mounted on the wall facing the area of interest (triangle in Fig. 5 left), we used an ellipse approximation. The room's operational area was roughly 3.6×4.8m, so each object's RSS map was 36×48px.

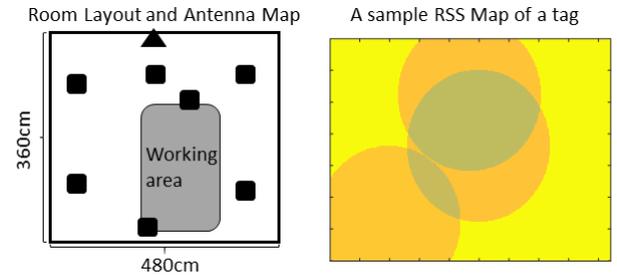

**Fig. 5.** Left: our RFID reader antenna configuration. Right: example RSS map visualization for a tagged object.

**Table 1. Our 25 RFID-tagged objects, including equipment sets.**

| Ophthalmoscope | BP Gauge Stand | Cardiac monitoring Adapter | Supply Pyxis | Small NRB |
|---|---|---|---|---|
| Otoscope | Pulse Ox Adapter | Stethoscope | Anesthesia Counter | Adult NRB |
| Blood Pressure Bulb | BP Cuff Inside | Breslow Tape | Counter | Small BVM |
| Thermometer | BP Cuff outside | Miami J Collar | Med Pyxis | Medium BVM |
| Bair Hugger Connector | Nasal Cannula | X-Ray Machine | Shears | Large BVM |

Generating the combined RSS map for all 25 types of tagged objects (Table 1) required two steps. First, we created maps for each object type (Fig. 5 right). Coverage areas of the 8 antennas were filled-in with that object's RSS (with zero values outside the coverage). The 8 resulting maps were then averaged, generating one RSS map per object type. Second, we created the combined RSS map by stacking the 25 2D maps for 25 object types into a 3D matrix.

### 4.3 Spatial Feature Extraction

We used the VGG net based structure for feature extraction [5]. Instead of vectorizing the extracted features and feeding them into fully-connected layers, we implemented a fully convolutional structure to better represent the features and conserve memory required for model training [26].

Because the different input matrices have differing dimensionality and sizes, and we faced hardware restrictions (limited GPU memory), we designed different ConvNets for each datatype based on the following rules (Fig. 6):

1. Use the maximum number of convolutional layers for which the hardware resources can train the networks within a reasonable time (around a week). It has been shown by several different sources that deeper networks generally perform better [4][18].
2. Although 2×2 max pooling has been used in many successful applications, the size of our audio-features map and RSS map could not be halved. We had to change the pooling sizes at certain pooling layers (Fig. 6, the pool 3 for RSS map) to shrink the activation map down to $1\times1\times N$ (the alternative to one dimensional fully-connected layers). The fully convolutional structure requires less resources for training and can be more representative for features compared with fully connected structure [26].

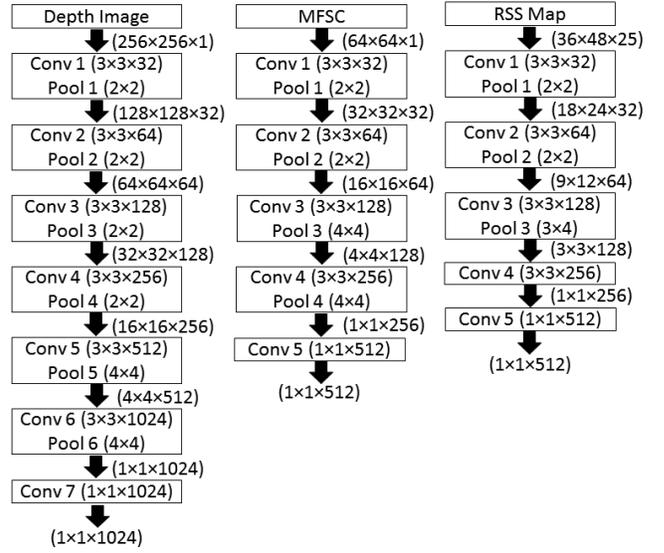

**Fig. 6. Our ConvNet structures for different sensors. Numbers of neurons in each layer are indicated.**

3. The size of one-dimensional convolutional layers (with dimension $1\times1\times N$) should be determined empirically. We determined the number of feature maps $N$ based on previous implementation [26].

The time window was often used in previous research to perform activity recognition on the data collected in a certain time period. We chose to not do so because activities in real-world scenarios often have very different durations, even activities of the same type (Fig. 3). Even with the slow fusion model [32], a time window arbitrarily limits the system's spatio-temporal features to the extent of the window. We addressed this problem by using a LSTM after each ConvNet to learn temporal associations between the spatial features over the timeline without using a time window (Fig. 4).

Note that, unlike previous research [22], we did not fuse the features from different datatypes immediately after the ConvNet modules. We used separate LSTMs to learn temporal associations for each sensor's data, and merged the resulting features later (Fig. 4).

### 4.4 Temporal Associations and Fusion

While ConvNets learn spatial features, LSTMs perform sequential learning [6]. Considering that the activity is a continuous concept, the feature sequences extracted by ConvNets should contain valuable temporal information. Complex and continuous activities, as in the trauma room, have varying durations. To avoid the exploding and vanishing gradient problem [33], we used a LSTM structure.

LSTM neurons inherit information over time using their "state" (or "memory"), which is independent from the input and output [33] (Fig. 7). We only implemented the forward LSTM (as opposed to a bidirectional one) because in activity recognition applications the future data is unknown. We provide a brief overview of LSTM to help the reader understand our structure and details can be found elsewhere [33]. Three major modules of a typical forward LSTM

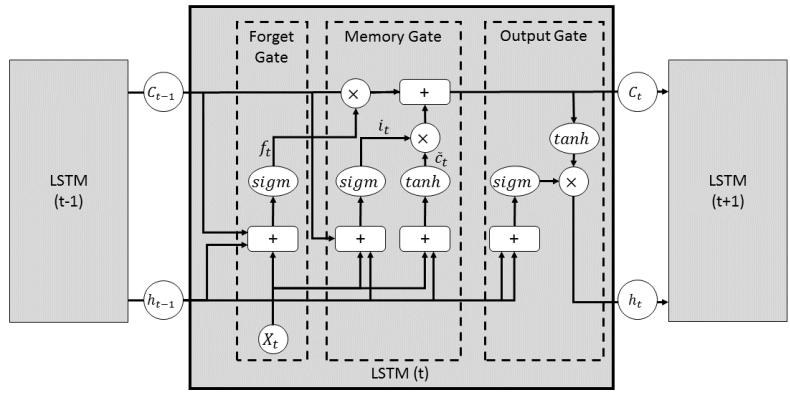

**Fig. 7. The structure of a single LSTM neuron. The current state *t* depends on the past state *t–1* of the same neuron.**

neuron are the forget gate, the memory gate, and the output (Fig. 7 dashed boxes).

The *forget gate* decides whether previous memories should be considered in the current time instance. For time instance *t*, we use $x_t$ to denote the neuron input, $C_t$ to denote its memory, and $h_t$ to denote its output. The output of forget gate at time *t* is:

$$f_t = \sigma(W_f[x_t, h_{t-1}] + b_f)$$

where $\sigma(\cdot)$ denotes the sigmoid activation function, $W_f$ denotes the weights, and $b_f$ denotes the bias.

The memory gate produces the current memory $C_t$ by generating a new candidate memory $\widetilde{C}_t$ and combining it with the old memory passed from the forget gate $f_t$:

$$i_t = \sigma(W_i[x_t, h_{t-1}] + b_i)$$
$$\widetilde{C}_t = tanh(W_c[x_t, h_{t-1}] + b_c)$$
$$C_t = i_t \cdot \widetilde{C}_t + f_t \cdot C_{t-1}$$

where $i_t$ is the input gate activation, $f_t$ is the forget gate activation, and *W* and *b* terms are the weights and biases.

Finally, the output gate decides the output of LSTM neuron:

$$o_t = \sigma(W_o[x_t, h_{t-1}] + b_o)$$
$$h_t = o_t * tanh(C_t)$$

where $o_t$ is the output gate activation, and $h_t$ is the output of the neuron at time *t*.

We designed a two-level LSTM structure to learn the temporal associations between features (Fig. 8). The *first level of our LSTM* is connected only backward to the associated convolutional layers. Each LSTM takes input from the last convolutional layer of its own datatype-specific ConvNet. The outputs of our three LSTMs (one for each data type) are then fully connected to a *fusion layer*. The fusion layer is a fully-connected layer that is partially backward-connected to each of the previous LSTMs and fully forward-connected to the following LSTM. An implementation of a similar fusion layer has shown that these layers can be tuned using normal backpropagation [22]. Finally, the second-level LSTM layer learns the temporal associations between merged features (output of the fusion layer).

We decided to add LSTM layers both before and after the fusion layer to first learn the temporal associations within the extracted feature of each data type, and then learn the temporal associations between data types. An alternative is to first merge the spatial information from each feature extractor (ConvNet structures) and then use LSTM to learn the temporal associations.

We also made the second LSTM half the size of the fusion layer for a smooth dimensionality reduction (no more than 10-times difference between adjacent layers) leading up to the final 35-activities class decision (Fig. 8).

### 4.5 Coding Layer

The last step in recognition of concurrent activities is to simultaneously make predictions for all activities in progress during the current time instance. The multiclass classifier (which predicts one label per time instance) is not applicable because it is incapable of making multiple positive predictions at once. We could have used multiple parallel binary classifiers (one per activity), but this would require an infeasible 35 fully-connected classifiers in our case, which is not scalable to larger numbers of activities.

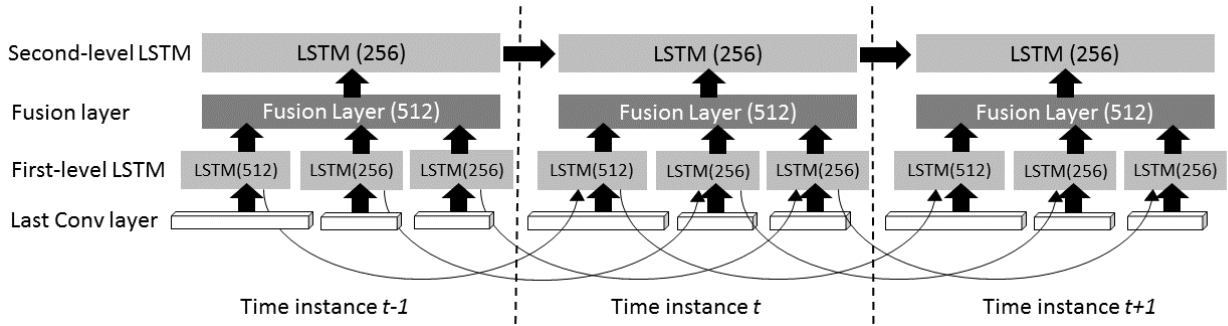

**Fig. 8. Our multilayer LSTM structure for temporal feature learning and feature fusion. (Detail from Fig. 4.) Arcs at the bottom symbolize the inherited state information across different time instances.**

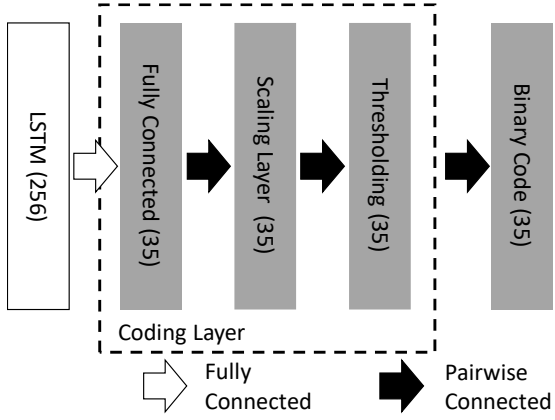

**Fig. 9. The structure of our coding layer. The numbers in the parentheses represent the number of neurons.**

We instead treated concurrent activity recognition as a coding problem in which a binary code represents the status of activities. In the code, a bit for each activity indicates whether it is in progress (one) or not (zero). We named this layer the *coding layer* (Fig. 9).

The coding layer has three components: the fully-connected layer, the scaling layer, and the thresholding layer (Fig. 9). The fully-connected layer takes the output of the previous LSTM layer and has $N$ neurons, where $N$ is the total number of activities ($N=35$ for the trauma resuscitation dataset). Depending on the activation function used in the fully-connected layer, a threshold (e.g. $\delta = 0.5$ for sigmoid activation) can be directly applied to the output of the fully-connected layer to generate the binary code. The weights of the fully-connected layer could be adjusted using regular backpropagation with the mean square error (MSE) loss function. This would face problems since backpropagation would try to minimize the MSE to reach the global optimum. However, since we still have to use a fixed threshold to turn the last layer's output into binary predictions, the global MSE optimum does not always lead to correct predictions.

We could either normalize the value range of neuron outputs in the last layer to fit a certain threshold, or we could train the threshold for each neuron in the last layer. We adopted the former approach by introducing a scaling layer with each neuron only pairwise-connected to neurons in the fully-connected layer (Fig. 9). We call neurons in the scaling layer "scaler neurons". The pairwise connections ensure that each threshold neuron will only be used for a single activity prediction. We used the sigmoid activation function for scaler neurons to scale the output from zero to one. Because the sigmoid neuron outputs values greater than 0.5 for a positive input and smaller than 0.5 for a negative input, we used the *leakyReLU* activation function [18] in the previous fully-connected layer to ensure the value for the scaling layer is distributed in both positive and negative axis. In this way, the minimum and maximum output from the threshold neuron is the same as the categorical label, and the global MSE optimum is equivalent to the categorical global optimum. The backpropagation error term is:

$$\Delta_i = -\gamma(o_i - g_i)f'(S_i)$$

where $\Delta i$ is the error term for backpropagation, $\gamma$ is the learning rate, $o_i$ is the $i^{th}$ output neuron, $g_i$ is the $i^{th}$ ground truth, and $f'(S_i)$ is the partial derivative of the $i^{th}$ sigmoid activation function. The outputs of the scaling layer are bounded between zero and one. A subsequent thresholding layer with the threshold $\delta = 0.5$ can be applied to generate the binary code for the prediction result.

## 5. IMPLEMENTATION
### 5.1 Model Training

We used Keras [34] with the TensorFlow as backend, and a GTX 1080 GPU with 8GB VRAM for training. We used adaptive moment estimation (Adam) optimization to adjust the learning rate, with decay rates $\beta_1=0.9$ and $\beta_2=0.999$ as suggested previously [35]. We also implemented dropout to prevent overfitting.

Most components of our network structure are available in deep learning frameworks, and can be implemented directly. We manually programmed our coding layer under the Keras framework.

During model training, we used random weight initialization; using the same initial weights or all-zeroes would lead to training problems. To make training more efficient, we delivered both positive and negative values to the coding layer by using the *leakyReLU* activation function in all convolutional layers[18] as the activation function in the convolutional layers. Because of different input data types and the structure of our convolutional network, the output scale in the last convolutional layer for different data types may be very different. Directly combining the features extracted from different sensor data in the later fusion layer would lead to low training efficiency. We used the sigmoid as the activation function for the last convolutional layer for each input to normalize the output of the activations to the range from negative one to one before data fusion.

Previous research [22] has shown that using partial data from process runs for both the training and testing causes overfitting, so we used whole independent runs for training versus testing. Specifically, we used 80% of our cases to train the system and used the remaining 20% of our cases for testing. Since we used LSTMs, the data within each case were not randomized during the training phase and the LSTM states were reset after each case in the training phase. In total, we performed 1,000 epochs of training.

### 5.2 Computer Memory Requirement Analysis

The memory analysis is an important step in the model design, especially for large and complex models. A good model should be trainable in a reasonable amount of time on commercially available computers. We estimated the memory required for feedforward (runtime) and backpropagation (training) procedures for our model as a reference for future implementations (Table 2). Note that our model requires significantly less memory compared to systems using multiple binary classifiers for concurrent prediction status (assuming each binary classifier uses a similar deep learning structure with softmax layers for final decision-making).

**Table 2. The required memory size (in bytes) in different layers of our deep learning structure during feedforward and backpropagation procedures for the data collected in each time instance.**

|  | Depth | | MFSC | | RSSMap | |
|---|---|---|---|---|---|---|
|  | **Feedforward** | **Backpropagation** | **Feedforward** | **Backpropagation** | **Feedforward** | **Backpropagation** |
| **Input** | 256*256*1*4=0.48M |  | 64*64*1*4=0.015M |  | 36*48*25*4=0.16M |  |
| **Conv1** | 256*256*32*4=8M | 3*3*32*1*4=0.001M | 64*64*32*4=0.5M | 3*3*32*1*4=0.001M | 36*48*32*4=0.21M | 3*3*25*32*4=0.025M |
| **Pool1** | 128*128*32*4=2M | 0 | 32*32*32*4=0.13M | 0 | 18*24*32*4=0.05M | 0 |
| **Conv2** | 128*128*64*4=4M | 3*3*32*64*4=0.064M | 32*32*64*4=0.25M | 3*3*32*64*4=0.064M | 18*24*64*4=0.1M | 3*3*32*64*4=0.064M |
| **Pool2** | 64*64*64*4=1M | 0 | 16*16*64*4=0.06M | 0 | 9*12*64*4=0.025M | 0 |
| **Conv3** | 64*64*128*4=2M | 3*3*64*128*4=0.256M | 16*16*128*4=0.12M | 3*3*64*128*4=0.256M | 9*12*128*4=0.05M | 3*3*64*128*4=0.256M |
| **Pool3** | 32*32*128*4=0.5M | 0 | 4*4*128*4=0.008M | 0 | 3*3*128*4=0.004 | 0 |
| **Conv4** | 32*32*256*4=1M | 3*3*128*256*4=1M | 4*4*256*4=0.016M | 3*3*128*256*4=1M | 3*3*256*4=0.008M | 3*3*128*256*4=1M |
| **Pool4** | 16*16*256*4=0.25M | 0 | 1*1*256*4=0.001M | 0 |  |  |
| **Conv5** | 16*16*512*4=0.5M | 3*3*256*512*4=4M | 1*1*512*4=0.002M | 1*1*256*512*4=0.5M | 1*1*512*4=0.002M | 1*1*256*512*4=0.5M |
| **Pool5** | 4*4*512*4=0.03M | 0 |  |  |  |  |
| **Conv6** | 4*4*1024*4=0.06M | 3*3*512*1024*4=16M |  |  |  |  |
| **Pool6** | 1*1*1024*4=0.003M | 0 |  |  |  |  |
| **Con7** | 1*1*1024*4=0.003M | 1*1*1024*1024*4=4M |  |  |  |  |
| **LSTM** | 512*4=0.002M | 1024*512*4*8=16M | 256*4=0.001M | 512*256*4*8=5M | 256*4=0.001M | 512*256*4*5=4M |
| **Fusion** | 512*4=0.002M | 512*512*4=1M |  | 512*256*4=0.5M |  | 512*256*4=0.5M |
|  | **Feedforward** | | | **Backpropagation** | | |
| **LSTM** | 256*5=0.001M | | | 512*256*4*8=4M | | |
| **Coding** | 35*2*4=0M | | | 256*35*4=0.03M | | |
| **Total** | 23M | | | 57.5M | | |

The design of deep learning structures (number of convolutional layers and neurons in fully-connected/LSTM layers, etc.) is empirical and based on experience, but is often determined by the hardware constraints. For every second of data, our design required around 23MB memory for the feed-forward prediction and ≈57.5MB for backpropagation training. This fact allowed us to train the system using mini-batches containing either one minute of data (for GPUs with a small 4GB memory) or two minutes (for high end GPUs such as GTX 1080). We assumed that ≈500MB of memory was occupied by the Linux OS. Our memory analysis also revealed that our system efficiently shared the multimodal CNN-LSTM feature extraction for the coding layer.

# 6. EXPERIMENTAL RESULTS

## 6.1 Trauma Resuscitation Dataset

### 6.1.1 Dataset Description
This dataset (synchronized RFID, depth, and audio data) was collected during 42 actual trauma resuscitations with over 30 hours of data [3]. The RFID data were collected with 8 antennas using 2 Impinj R420 readers; the depth video and audio data were collected with the Kinect sensor (the RGB camera was not used). The ground truth coding was done manually by medical experts from surveillance videos.

We selected 35 activities (Table 3) for this study. Because the performed activities depend on various factors including patient attributes, the dataset contains varying amounts of data for different activities. Trauma resuscitation is inherently a fast-paced teamwork process and concurrent activities are common. In our dataset, about 50% of the time instances contained concurrency (Fig. 1(a)).

### 6.1.2 Experimental Results
We first evaluated the system by calculating the average accuracy for the recognition of concurrent activities:

$$acc = \frac{\sum_{i=1}^{A}\sum_{j=1}^{T} XNOR(P_{ij}, G_{ij})}{A \cdot T}$$

where $A$ is the total number of activities ($A$=35 in our case) and $T$ is the total number of time instances in the testing set. $XNOR(\cdot)$ denotes the XNOR operation. $P_{ij}$ denotes the prediction results for activity $i$ at time instance $j$. Similarly, $G_{ij}$ denotes the ground truth for activity $i$ at time instance $j$. The average accuracy for 35 activities was 93.48% (Fig. 10).

As mentioned, the positive (activity is in progress) and negative (activity not in progress) samples in our dataset were imbalanced, and therefore the accuracy alone does not reflect well the system performance. We included the F1-measure, Informedness and Markedness [36] to better evaluate the system performance (Fig. 10).

**Table 3. The 35 activities and their codes used in this study. Shaded are activities that used tagged objects (Table 1).**

| Code | Activity | Code | Activity | Code | Activity |
|---|---|---|---|---|---|
| A | Abdomen Assessment | EA | Exposure Assessment | M | Mouth Assessment |
| AA | Airway Assessment | EAR | Ear Assessment | N | Nose Assessment |
| B | Bolus Delivery | EC | Exposure Control | NE | Neck Assessment |
| BA | Breathing Assessment | ET/TT | Confirm ET/Trach Tube Placement | OS | Oxygen Saturation |
| BC | Breathing Control | EY | Eye Assessment | PC | Pulse Check |
| BK | Back Assessment | F | Face Assessment | PE | Pelvis Assessment |
| BP | Blood Pressure | G | Genital Assessment | PU | Pupil Assessment |
| C | Chest Assessment | GCS | GCS Calculation | RLE | Right Lower Extremity Assessment |
| CA | Circulation Assessment | H | Head Assessment | RO | Relieve Obstruction |
| CC | Circulation Control | I | Intubation | RUE | Right Upper Extremity Assessment |
| CR | CPR | LLE | Left Lower Extremity Assessment | SA | Secondary Survey Adjuncts |
| CS | C-Spine Stabilization | LUE | Left Upper Extremity Assessment | | |

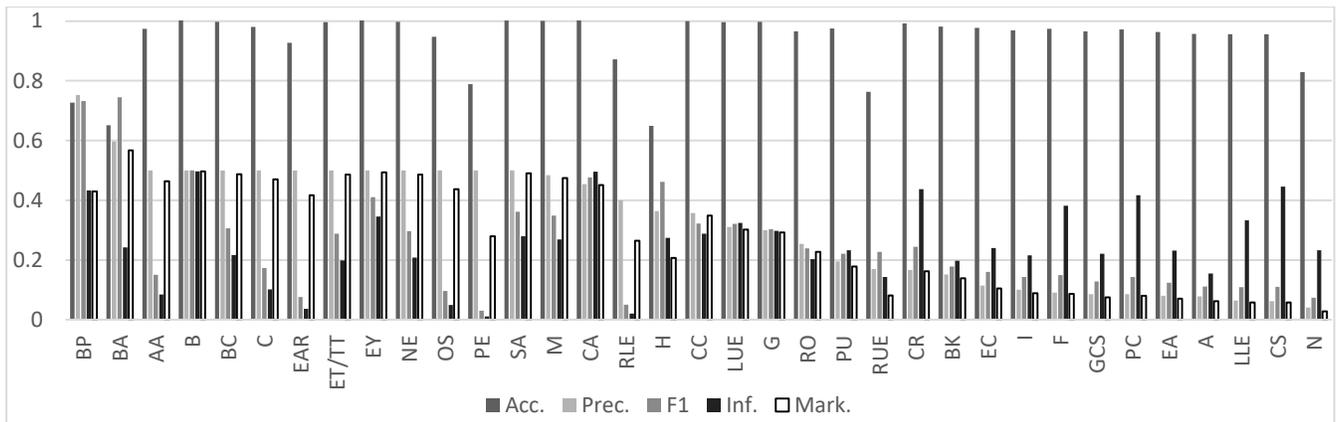

**Fig. 10. Our system evaluation with accuracy (Acc.), precision (Prec.) F1 score (F1), informedness (Inf.) and markedness (Mark.). Sorted in the descending order of precision.**

We also analyzed the activity recognition performance for 3 types of activities as case studies:

1. The system achieved stable and accurate prediction performance for activities with long durations and activities that usually occurred simultaneously, such as BC and CA (Fig. 2; acronyms in Table 3). This likely occurred because the LSTM was able to learn the temporal associations between simultaneously occurring activities. Even if the ConvNet failed to extract all the spatial features, the LSTM was still able to use temporal associations between features to make correct predictions.
2. The activities where people had very distinct postures or used RFID-tagged objects generally had better recognition accuracy. For example, when performing BA, the provider needs to stand in the patient head area and slightly lean over the patient. This unique posture can be effectively learned by the ConvNet. The RFID and microphone array provided useful information for activities that required object-use. For example, blood pressure cuffs and bulbs tagged with passive RFID tags provided useful information to detect the BP activity.
3. Rare or short activities (such as CR, I in Fig. 3) were hard to recognize for two reasons. The dataset usually had not enough training samples for rare or short activities for the ConvNet to learn the representative features. Therefore, the network was not sensitive enough to such activities. Also, rare activities did not occur in every case, providing only weak temporal associations for the LSTM to learn.
4. The activities that did not use any objects, did not have unique performer's posture or could not be clearly captured by the camera, were hard to detect (e.g. LLE and PC). These problems were generally caused by limitations of our sensors. A potential solution is to install additional depth sensors on the wall opposite of the current Kinect and use other sensors to better detect human motion.

Finally, we compared our system to previous systems for medical activity recognition [3][10] and process phase detection [22][37][38] (Table 4). Given only a few published works, we also compared our system with activity recognition in other domains [32][39] (Table 4). Because different systems were proposed for different application scenarios, with different sensors and activities of different complexity (staged vs. real vs. concurrent real activities), we cannot declare any system as the best. We found that:

1. Generally, large-scale activity recognition is more challenging than small scale one. A work on video classification [32] achieved a fairly low accuracy (top 1 accuracy) using ≈500 labels, but their system is still one of the best for activity recognition or video classification.

**Table 4. Performance comparison of activity recognition systems.**

| Process phase detection system | Accuracy | #Activities | Sensor | Concurrent activity recognition |
|---|---|---|---|---|
| RFID based activity recognition system [10] | 75% | 10 activities | Passive RFID only | Yes |
| Deep learning for RFID-based activity recognition [3] | 80% | 10 activities | Passive RFID only | No |
| Online resuscitation phase detection [22] | 80% | 6 resuscitation phases | Passive RFID and depth sensor | No |
| Modeling and online recognition of surgical phases using hidden Markov models [37] | 86% | 14 surgical phases | Signal from medical equipment | No |
| Phase recognition during surgical procedures using embedded and body-worn sensors [38] | 77% | 7 surgical activities | Wearable sensor and RFID | No |
| Large scale video classification with ConvNet [32] | 64% (top 1) 82% (top 5) | 487 activities | RGB camera | No |
| Recurrent neural networks for analyzing relations in group activity recognition [39] | 80% | 10 activities | RGB camera | No |
| Proposed system | 93.9% (0.32 mAP) | 35 activities | Passive RFID, depth sensor and microphone array | yes |

2. Sensor data quality significantly influences the activity prediction performance. The tradeoff is that putting sensors (cameras, RFID antennas) closer to the scene will lead to cleaner data, but the system might interfere with people's work [38]. Putting the sensors away from workers will avoid interference with work, but the camera's view occlusion and the RFID's radio noise from people movement will impair the system performance. Our priority was to avoid interference with work and then try to maximize the performance.

3. Concurrent activities are frequent in real-world scenarios and in large datasets video clips may be labeled with multiple tags [32]. Most previous research got around the multiclass classifier's limitations by simply duplicating the data with concurrent activities and labeling them with the labels of those activities. Our experimental results showed that our system is able to achieve similar recognition performance for concurrent activities, using a single classifier instead of multiple binary classifiers.

## 6.2 Charades Dataset

To evaluate our system on concurrent activity recognition using different data sources, we selected the Charades dataset, containing 157 action labels [11]. We selected this dataset for two reasons. First, it contains well-labeled daily living activity videos. Since our trauma resuscitation experiment did not use an RGB camera, we also wanted to demonstrate the system's applicability to other sensors. Second, concurrent activities are common in daily living scenarios and this dataset demonstrates our system's ability to recognize concurrent activities.

To make our network structure work with the dataset containing RGB video only (480×360px) and meet the hardware limitations, we downsampled the videos to 256× 256px. We modified our proposed network structure to use a single ConvNet and removed the fusion layer to work with RGB input only. Since there were 157 activities, the coding layer had 157 neurons, and generated 157-bit binary codes every second, representing the respective activity statuses.

We compared our results with results reported in other research [11], using their suggested training and testing data split. We used the GTX 1080 GPU to train the model, which took about 10 days to converge, but the network could be further tuned. Given the predictions, we calculated the mean average precision (mAP) to compare our system performance with other approaches [11] (Table 5).

Note that the previous classification evaluation [11] did not describe how they handled the concurrency, so we assumed they trained multiple one-vs-rest SVMs for each activity. Our comparison showed that our system achieved the performance competitive with the baselines (Table 5). Our network structure can be easily implemented with other data sources under scenarios with or without concurrent activities. Considering that we only used the low-resolution videos for system training and we only trained the system for a limited time, there is still opportunity for fine tuning the system to achieve better performance.

## 6.3 Olympic Sports Dataset

Our system also works well with regular single-activity prediction from single sensor data and we used the Olympic sports dataset [12] as a demonstration. The dataset contains videos of 16 different sports (Table 6) with only RGB frames and no audio. All videos were on YouTube and class labels were annotated with the help of Amazon Mechanical Turk.

Similar to the Charades dataset, we slightly modified our network structure to have only an RGB-video ConvNet branch and removed the fusion layer. As suggested by previous research that used the same dataset, 80% of data in each label was used for training and the remaining 20% for testing and cross validation. We calculated the average precision to compare our system performance with other approaches that used the same dataset [12][40] (Table 6).

**Table 5. Mean average precision (mAP, %) for 157-action classification using results reported in [11] and our approach.**

| Approach | mAP |
|---|---|
| Random | 5.9 |
| C3D | 10.9 |
| AlexNet | 11.3 |
| Two-Stream_B | 11.9 |
| Two-Stream | 14.3 |
| IDT | 17.2 |
| **Our System** | **12.4** |

The comparison showed that our approach achieved the best performance in 4 out of 16 activities. Existing research did not mention their specific split of training and testing data, so their performance might be slightly different for the same dataset. The network tuning became slower with smaller learning rates, so we stopped the model training at 150 epochs due to time limitations. Our results showed that our system could be easily implemented for multiclass classification problems as well as use single or multiple sensors, which met our design goal. The system performance could be improved with more convolutional layers and deep LSTM layers, or using the existing multimodal structure for video classification [32]. Further training the system with lower learning rates for longer times would certainly improve the performance. However, tuning the network for specific applications is beyond the scope of this study.

## 7. DISCUSSION
### 7.1 The Number of Activity Labels Matters

There were 35 activity types labeled in the trauma resuscitation dataset and 157 in the Charades dataset. Depending on the application, different types of activities can be labeled for tracking, while other activities will be ignored. In addition, activities could be differentiated or generalized into finer or broader categories. Both approaches may significantly increase or decrease the total number of activities in the dataset, changing the extent of co-occurring activities or even creating false "idle periods". This outcome may, in turn, influence the recognition performance. For example, fewer activity labels will likely result in fewer concurrent activities, reducing the volume of temporal relationships for the LSTM to learn.

We applied our system to the well-known CIFAR 100 dataset, containing 100 types of images to perform concurrent image recognition. We chose this approach as equivalent to concurrent activity recognition with multi-sensory input [42], because activity-recognition model training with video input takes much longer time. To simulate concurrent image recognition (recognizing multiple targets in a single image at once), we randomly selected 6 images from the CIFAR 100 dataset and combined them into a single large image (Fig. 11). We thus generated 50,000 training images and 10,000 testing images. We trained and tested with different numbers of labels ranging from 10 (only 10 image types were labeled, and the others were ignored) to 100 (all images labeled). We calculated the accuracy and average precision (AP) for different scenarios (Fig. 12). The results showed that the system trained and tested on a larger number of labels will likely have higher testing accuracy and lower AP. This result may be because with a larger set of labels and a few positive labels in each subset, accuracy will be less representative. The total number in the denominator increases and the number of correct predictions often stays roughly the same. The AP tends to be lower, since with more activities, it is easier for the system to confuse one activity with another and have false alarms.

**Table 6. The activity recognition average precision values for Olympic Sports Dataset.**

| Code | Our method | Method in [41] | Method in [40] |
|---|---|---|---|
| Basketball layup | 79.83% | 82.1% | **85.5%** |
| Bowling | 64.05% | 53.0% | **64.3%** |
| Clean and jerk | **80.63%** | 70.6% | 78.2% |
| Discus throw | 41.05% | 47.3% | **48.9%** |
| Diving platform 10m | 51.28% | **95.4%** | 93.7% |
| Diving springboard 3m | 49.40% | **84.3%** | 79.3% |
| Hammer throw | **72.42%** | 71.2% | 70.5% |
| High jump | **32.52%** | 27.0% | 18.4 % |
| Javelin throw | 58.39% | **85.0%** | 79.5% |
| Long jump | 41.03% | 71.7% | **81.8%** |
| Pole vault | 59.98% | **90.8%** | 84.9% |
| Shot put | 21.46% | 37.3% | **43.3%** |
| snatch | 46.00% | 54.2% | **88.6%** |
| Tennis serve | 11.06% | 33.4% | **49.6%** |
| Triple jump | **39.50%** | 10.1% | 16.1% |
| vault | 61.54% | **86.1%** | 85.7% |
| Average | 49.91% | 62.5% | **66.8%** |

Research or real-world applications of activity recognition will inevitably track only a subset of people's activities and non-tracked activities will decrease temporal relationships or appear as "idle times". As our experiment shows, using only the accuracy or AP cannot fully reveal the system performance. Therefore, system evaluation results should report as well the distribution of co-occurring activities and the fraction of idle time (Fig. 1).

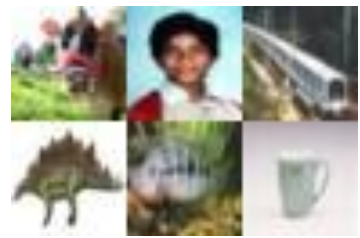

**Fig. 11. An example of combined images for multi-target or "concurrent" recognition.**

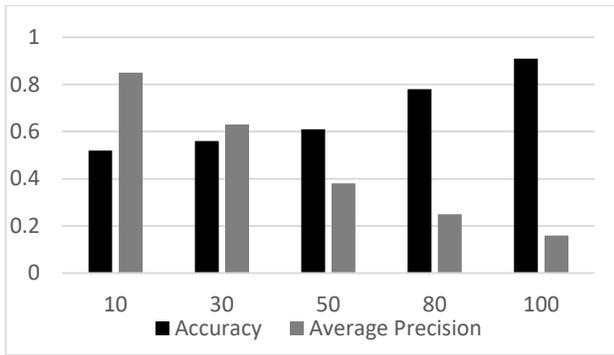

Fig. 12. The comparison of accuracy and mean average precision for different number of labels.

## 7.2 System Limitations

**From data quantity to quality:** A key challenge we had for activity recognition was insufficient data. Experimental results show that rare or short activities are harder to recognize due to limited number of training samples [10]. In addition, the data recorded by a single sensor type also limited the type of activities to be tracked (e.g. the RFID sensor alone cannot effectively detect activities that do not use taggable objects). To mitigate the insufficient data issue, we used different sensors and a multimodal deep learning structure for activity recognition.

Even with sufficient data and multiple sensors, there may still be limitations caused by the data quality because of sensor positioning. For example, depth sensors were installed on the sidewalls of the trauma room to maximize the view range, but there was still view occlusion caused by people blocking. View occlusion leads to information loss resulting in failure to predict certain activities. Other sensors also suffer data quality issues: a long distance between the microphones and speakers leads to poor speech quality and people movement leads to RF signal fading.

Optimal number of sensors and their positioning have not been fully studied and will be part of our future work. Currently we only used the RSS data from RFID tags and including the phase angle and Doppler shift will provide additional information.

**Generalization:** The generalization is another limitation of the current system. In image classification [5], the main target remains similar even in images taken by different cameras at different distances. The RFID data recorded by different antenna settings (different antenna position, different number of antenna, or different tagging strategy) will be different. This makes the proposed system only generalize well with similar hardware configurations. The same generalization problem will exist for other mobile sensors, such as wearable sensors, which require each sensor node to be worn on a specific body part. A potential solution is to use reference tags previously used for people tracking and localization. Instead of using the RSS directly measured by the antennas, we can try to measure the relative RSS between tags and certain reference tags. In this way, though the absolute RSS value will be different under different antenna configurations, the scaled relative RSS values should remain similar.

**Ground truth coding efficiency:** The current model training relies on manually generated ground truth, which is laborious and requires domain expertize. Training deep learning models will require a large volume of ground truth coding and for some domains. Instead of relying on human coding for every second of data, we believe a semi-supervised model will help save coding time and make ground truth coding more efficient [43].

## 7.3 Future Extensions

**Applications beyond activity recognition:** Due to the ability to treat the task as coding problem, the proposed system can be applied to many other fields in activity recognition. For example, if we treat the general image/video classification as coding problems, we can directly apply the proposed system for image and video classification which should work similarly as multiclass classification models using softmax layers for decision-making. However, if a video or video contains multiple targets and requires image captioning or description, the multiclass classification won't be able to make concurrent predictions at once. The common solution for image captioning and description is to first search for the region of interests using region proposal method [44] and then run the recognition for targets in each region [45]. Based on the proposed system structure, an alternative is to consider the image captioning as a coding problem, where each digit in the binary code denotes a possible target. With the proposed model, the image with multiple targets can be labeled without using the region proposal with a single neural network.

We performed some preliminary experiments using the famous MNIST dataset [46] and CIFAR 100 dataset [42] for multi-target image recognition. We randomly selected 6 images form each dataset and put them together as one large image. The preliminary results show 94.1% accuracy and 0.96 mAP for multiple digits recognition in MNIST dataset and 91.3% accuracy with 0.16 mAP for CIFAR 100 dataset.

**Daily activity recording:** The proposed system can be used for real-life complex activity recognition with multisensory data. Since wearable sensors and mobile phones have multiple built-in sensors, such as light sensors, gyroscopes, microphones, etc., the proposed system can be used with these mobile devices for daily living monitoring and recording. There are many current studies focused on daily living monitoring with mobile sensors, but most of them only trained and tested for a few activities [47]. In such cases, the system is not guaranteed to achieve the same performance compared to testing sets with limited activities. Our system is able to learn the spatial-temporal features from the recorded raw data and make large scale activity recognition even with concurrent activities.

## 8. CONCLUSION

We presented a system structure that uses deep learning for concurrent activity recognition from multiple sensory datatypes. We tested our system with several different datasets and with one general system achieved results comparable to existing specialized systems. We believe our research delivers the following benefits to the community:

1. A feature-learning structure capable of learning the spatial-temporal features directly from the raw data via ConvNet and LSTM structures, which requires neither manual feature selection nor classifier selection.
2. The fast and resource-efficient coding-layer structure for concurrent activity and multi-target recognition.
3. A demonstration of the RSS-map data representation for RFID data. Our RSS map works well with ConvNets and shallow classifiers. It can be used for other sensor types, such as WIFI nodes, Bluetooth nodes, or wearable sensors, as well.
4. Detailed memory analysis for each layer of our deep multimodal structure, which can be used as a reference for future research.
5. Experimental results using published datasets such as the Charades and Olympic datasets, which can be used as a baseline for future research.
6. Examples of extending our framework to other uses such as data classification, captioning and description.